\documentclass[11pt,a4paper]{article}
\usepackage{emnlp2020}
\usepackage{times}
\usepackage{latexsym}
\usepackage{amssymb}
\usepackage{amsmath}
\usepackage{graphicx}
\usepackage{booktabs}
\usepackage{mathrsfs}
\usepackage{makecell}
\usepackage{graphics}
\usepackage{subfigure}
\usepackage{ulem}
\usepackage{microtype}

\aclfinalcopy 
\def\aclpaperid{3012} 

\title{Unsupervised Reference-Free Summary Quality Evaluation via Contrastive Learning}

\author{Hanlu Wu$^1$\thanks{\ \ Equal contribution}, Tengfei Ma$^2$\footnotemark[1], Lingfei Wu$^2$, Tariro Manyumwa$^1$, Shouling Ji$^1$\\
$^1$Zhejiang University,
$^2$ IBM Research\\
  \texttt{wuhanlu@zju.edu.cn, tengfei.ma1@ibm.com,wuli@us.ibm.com},\\
  \texttt{tmanyumwa@yahoo.com, sji@zju.edu.cn} }

\date{}

\begin{document}
\maketitle
\begin{abstract}
Evaluation of a document summarization system has been a critical factor to impact the success of the summarization task. Previous approaches, such as ROUGE, mainly consider the informativeness of the assessed summary and require human-generated references for each test summary. In this work, we propose to evaluate the summary qualities without reference summaries by unsupervised contrastive learning. Specifically, we design a new metric which covers both linguistic qualities and semantic informativeness based on BERT. To learn the metric, for each summary, we construct different types of negative samples with respect to different aspects of the summary qualities, and train our model with a ranking loss. Experiments on Newsroom and CNN/Daily Mail demonstrate that our new evaluation method outperforms other metrics even without reference summaries. Furthermore, we show that our method is general and transferable across datasets.

\end{abstract}

\section{Introduction}
Recently, there has been great success in automatic text summarization and generation~\cite{huang2020knowledge,leclair2020improved,chen2019reinforcement}. To better compare and improve the performance of models, evaluation for such systems has been a problem of interest. The selection of evaluation metrics will greatly affect the assessed quality of a generated summary and thus affect the evaluation of summarization models. 

The most ideal metric is definitely human judgement, which is often treated as the gold standard. But human evaluation is time-consuming and labor-intensive, an automatic evaluation metric that cannot only save human resources but also simulate the ability of human judgement is of crucial importance. 

Most of the existing automatic evaluation methods assess a summary by comparing it with reference texts written by humans. Some of them are model-free and simply use hand-crafted matching functions to calculate the similarity between the candidate summary and the reference  \cite{papineni2002bleu,lin2004looking,banerjee2005meteor}. These methods consider both the reference and the candidate as a sequence of tokens or n-gram blocks. For instance, as the de facto standard evaluation metric, ROUGE \cite{lin2004looking} calculates the n-gram overlap between the machine-generated summaries and reference summaries. Although these methods have the advantage of interpretability and efficiency, they are found to correlate poorly with human evaluation~\cite{Novikova2017Why}. 

To reduce the requirement of exact word matching, some recent work tried to match the reference and the candidate summary in the embedding space of words or sentences \cite{zhang2019bertscore,clarketal2019sentence,zhao2019moverscore}. For instance, BERTScore \cite{zhang2019bertscore} uses contextual word embeddings generated by BERT and performs a greedy matching to obtain the maximum cosine similarity between two texts.
These methods are proved to correlate better with human judgement than ROUGE on many datasets, which demonstrates the effectiveness of using contextual embeddings.

\begin{table*}[ht]
\centering

\scalebox{0.85}{
\begin{tabular}{lccc} 
\toprule

    \textbf{ } & \textbf{Semantic} & \textbf{Linguistic}  & \textbf{Else} \\
    \cmidrule(l){1-4}
    
   \makecell[l]{~DUC-05, DUC- 06 and DUC-07\\ ~\cite{Xenouleas2019SumQE}}
 & \makecell[c]{focus, \\non redundancy}  & \makecell[c]{grammaticality,\\structure \& coherence}  & referential clarity \\
   
   \cmidrule(l){1-4}
    ~Newsroom 60~\cite{sun2019feasibility} & \makecell[c]{relevancy, \\informativeness,\\ unnecessary content,\\verbosity}
  & -  & \makecell[c]{perfect surrogate,\\ continue reading}  \\
  
  \cmidrule(l){1-4}
    *CNN/Daily Mail~\cite{chaganty2018price} &-    & \makecell[c]{fluency,\\overall quality,\\ redundancy}  &-   \\
    \cmidrule(l){1-4}
    *Newsroom~\cite{grusky2018newsroom} & \makecell[c]{informativeness, \\relevancy}   & \makecell[c]{coherence,\\fluency}  &-   \\
    \cmidrule(l){1-4}
   \makecell[l]{~NYT and CNN/Daily Mail \\~\cite{sharma2019entity}} & informativeness   & \makecell[c]{grammaticality,\\ coherence}  &-  \\

\bottomrule
\end{tabular}
}
\caption{Evaluation Dimensions of Different Summarization Datasets. *: the dataset is used in our experiments. Note that for the dataset proposed by \citeauthor{chaganty2018price} (\citeyear{chaganty2018price}), all the three dimensions focus on evaluating the linguistic quality of summaries.}
\label{tab:eval-dimensions}
\end{table*}

However, the aforementioned methods all have some intrinsic drawbacks: these methods always need at least one human-generated reference to assess a candidate summary. References written by humans are costly to obtain. In addition, most of them only consider the semantic similarities with references, i.e. semantic qualities of the summaries, which ignores the linguistic qualities and other important aspects. In this paper, we propose a new unsupervised contrastive learning framework for automatically evaluating the summary qualities without comparing with reference summaries or training with human ratings. Specifically, we design an evaluator to consider both linguistic and semantic aspects of a summary. Then for each of the aspect we create a set of negative samples by perturbing the training samples. We compare the scores of original training samples and the negative samples to obtain the contrastive loss function and learn the evaluator. The experiments on Newsroom and CNN/Daily Mail demonstrate that our new evaluation method has much higher correlation with human judgement.

We summarize our contributions as follows:
\begin{itemize}
    \item We develop a new unsupervised method for summary quality evaluation which considers both linguistic and semantic aspects.
    \item We creatively make negative samples with respect to our evaluation metric and train the evaluator by contrastive learning.
    \item Our evaluator requires no reference summaries or human ratings but achieves the best performance on single-document summarization datasets, and the trained evaluator can be easily used across different datasets.
\end{itemize}

\section{Related Work}

\subsection{Existing Evaluation Metrics}

\subsubsection{Reference-based Metrics}
Most of the existing automatic metrics for summarization evaluation assess a model-generated summary (i.e. the candidate) by comparing it with a human-authored summary (i.e. the reference).

Some metrics are model-free and their scoring basis are often easy to interpret~\cite{papineni2002bleu,lin2004looking,banerjee2005meteor}. For instance, as the most widely used metric for summarization evaluation, ROUGE~\cite{lin2004looking} measures the co-occurrence of n-grams or substrings between the reference and the candidate.

Most of the model-based methods~\cite{zhang2019bertscore,zhao2019moverscore,clarketal2019sentence} compare the embeddings of the reference and the candidate. BERTSCore~\cite{zhang2019bertscore} uses pretrained BERT contextual embeddings~\cite{devlin2019bert} and performs a greedy matching to obtain the maximum cosine similarity between embeddings of tokens in the two texts. \citeauthor{clarketal2019sentence} (\citeyear{clarketal2019sentence}) proposed metrics based on sentence mover's similarity (SMS) by leveraging sentence-level embeddings for evaluating multi-sentence texts. MoverScore~\cite{zhao2019moverscore} combines 
n-gram contextual embeddings and Earth Mover's Distance. BERTScore can be viewed as a special case of MoverScore. NUBIA \cite{Kan2020NUBIANB} considers three aspects of features of the reference-candidate pairs and aggregates the extracted features using a neural network regressor.

These metrics have a common drawback that the evaluation is based on costly human-authored references.  To assess the quality of a generated text summary, we need to obtain a corresponding ground-truth reference.

\subsubsection{Reference-free Metrics}
Some work discussed how to evaluate the quality of generated text in the reference-free setting~\cite{Louis2013Automatically,peyrard2017learning,peyrardgurevych2018objective,shimanaka2018ruse,Xenouleas2019SumQE,sun2019feasibility,bohm2019better,chen2018A,gao2020supert}. \citeauthor{Louis2013Automatically} (\citeyear{Louis2013Automatically}), \citeauthor{peyrard2017learning} (\citeyear{peyrard2017learning}) and \citeauthor{peyrardgurevych2018objective} (\citeyear{peyrardgurevych2018objective}) leveraged regression models to fit human judgement. RUSE \cite{shimanaka2018ruse} use sentence embeddings generated by three different models and aggregate them using a MLP regressor. \citeauthor{Xenouleas2019SumQE} (\citeyear{Xenouleas2019SumQE}) proposed a method that also uses a regression model to predict the scores, while the predictions are based on hidden representations generated using BERT~\cite{devlin2019bert} as the encoder. However, these methods require ratings assigned by human annotators as training data which are also costly to obtain. In contrast, our method is \emph{unsupervised} and requires no human ratings for training.

\citeauthor{sun2019feasibility} (\citeyear{sun2019feasibility}) discussed both reference-based and reference-free settings for summarization evaluation. Their method basically converts both the generated text and the text for comparison (denoted as \textit{T}) into hidden representations using encoders like ELMo~\cite{Peters2018Deep} and calculates the cosine similarity between them, \textit{T} in the reference-based setting and the reference-free setting stands for the human-authored reference text and the source document text, respectively. However, the experiment results show that their method's correlation with human ratings is lower than ROUGE, especially in the reference-free setting.~\citeauthor{chen2018A} (\citeyear{chen2018A}) designed a  Question-Answering based method to compare the content difference of two texts. Although this method provides a novel perspective and the evaluation basis is easy to interpret, the results show that it has not achieved better performance than ROUGE considering the lower correlation with human ratings.

SUPERT generates pseudo references and evaluates the quality of the test summaries by calculating word mover's distance between the pseudo reference summaries and the test summaries~\cite{gao2020supert}. It is similar to MoverScore~\cite{zhao2019moverscore} which uses the human-authored references instead of pseudo references. However, SUPERT mainly focuses on multi-document summarization evaluation, and its performance is inevitably worse than MoverScore. 

The work closest to our model is an evaluation method for natural language generation (NLG) systems proposed by \citeauthor{zhou2020learning} (\citeyear{zhou2020learning}). They implemented the sample-level evaluation by comparing a pair of texts. However, their method requires a set of different NLG systems and they need to generate weak supervision sample pairs from different checkpoints of a system. For testing, they also need to compare different samples to obtain a comparison score. In contrast, our model focuses on summarization evaluation; we do not need generated texts from many systems and different checkpoints of a system: all our negative samples are created by modifying the existing summaries; and in the test phase no comparison between different summaries is needed.

\subsection{Dimensions of Evaluation}

We investigated a few summarization datasets. As shown in Table~\ref{tab:eval-dimensions}, different datasets consider different evaluation dimensions. We observed that these dimensions can be roughly divided into three classes: the semantic quality (Semantic), the linguistic quality (Linguistic), and other dimensions that can be hardly classified (Else). In this paper, we design our method to cover both dimensions of semantic quality and linguistic quality.

\begin{figure}[htbp]
\centering
\begin{minipage}[t]{\linewidth}
    \centering
    \includegraphics[width=0.6\linewidth]{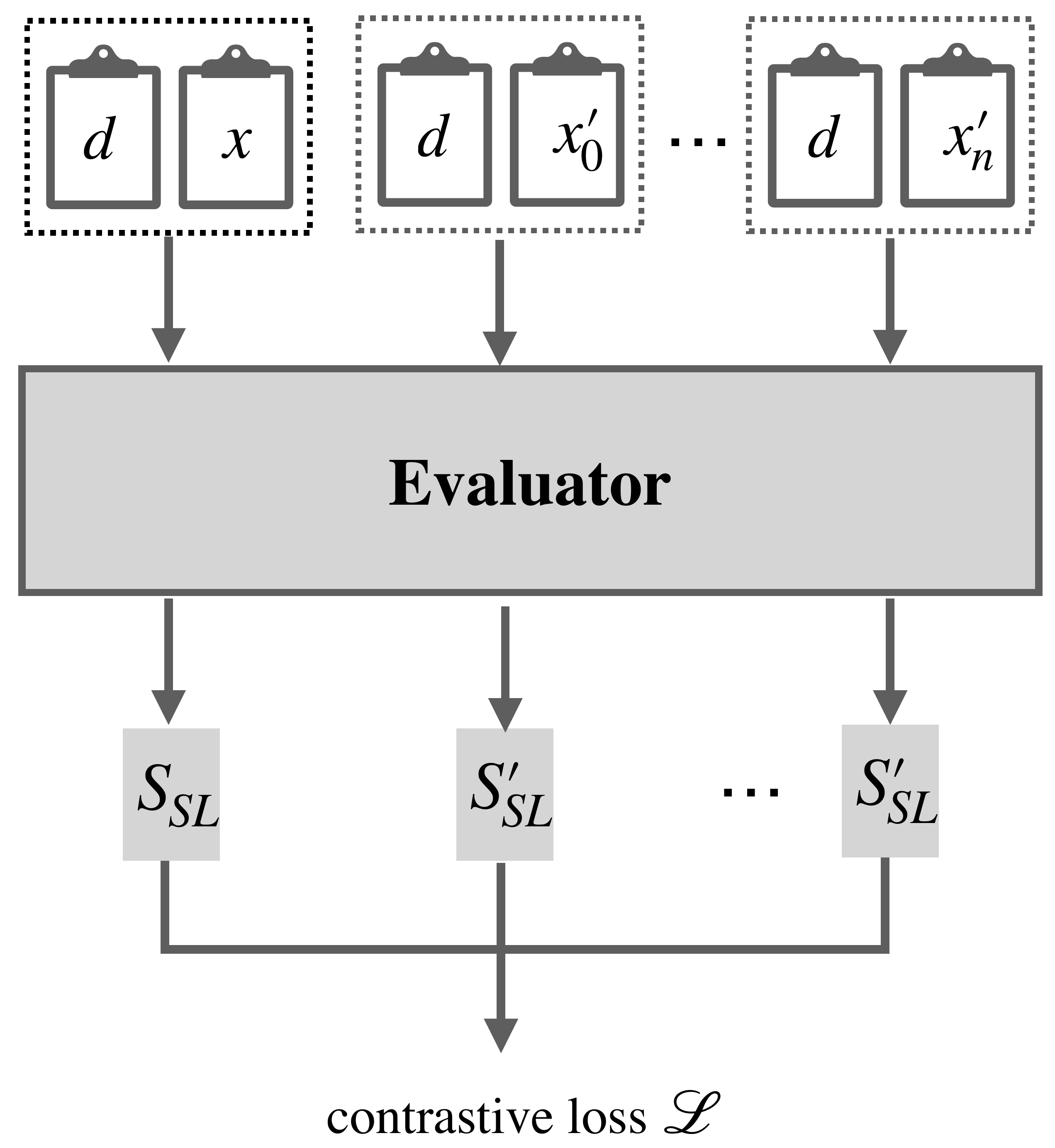}\\
    \includegraphics[width=0.9\linewidth]{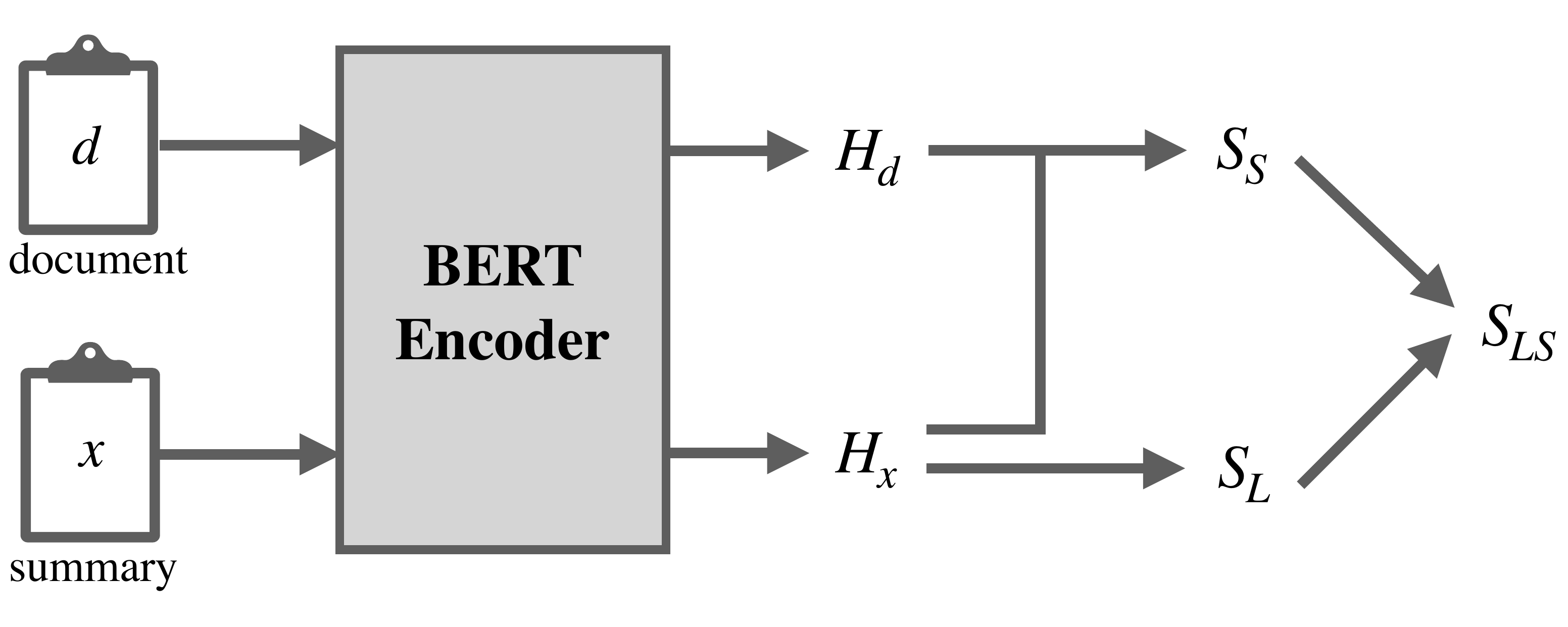}\\
\end{minipage}%
\caption{Model Framework. The top figure describes the framework for contrastive learning, where for each document $x$, we create different types of negative samples and compare them with $x$ to get a ranking loss. The bottom figure is the evaluator which generates the final evaluation score. For short, here we use $S_S$, $S_L$ and $S_{LS}$ to indicate $S\_Score$, $L\_Score$ and $LS\_Score$.}
\label{fig:evaluaator}
\end{figure}

\section{Method}

As shown in the previous section, two of the most important factors that impact the summary qualities are linguistic quality and semantic quality. Linguistic quality indicates how natural the generated summary is; it generally includes the fluency of each sentence, the coherence of entities/consecutive sentences, and the correctness of grammars. Semantic quality indicates whether a summary expresses the most important information of the original documents; it generally includes informativeness, relevance, and redundancy, etc. We consider both aspects and design our method in the following sections. Our model architecture is shown in Figure~\ref{fig:evaluaator}. The figure contains two parts, first we design our evaluator to assign scores to summaries based on a BERT encoder. Then we create negative samples and use a contrastive learning framework to train the evaluator.

\subsection{Evaluating Semantic Quality}
\label{section:evaluate-semantic-quality}

To better evaluate the semantic quality, we utilize the contextualized embeddings of BERT~\cite{devlin2019bert}. BERT takes in a sequence which always starts with a special classification token \texttt{[CLS]} as input, and outputs the representation of this sequence. Each token has its own hidden state. The hidden state corresponding to \texttt{[CLS]} is supposed to aggregate information from the whole sequence. We design our evaluation model as follows.

Formally, let $S_x$ and $S_d$ be the sequence of tokens in the summary $x$ and the source document $d$ from which $x$ is generated. A sequence of tokens is encoded into a sequence of token embeddings $H$ by the BERT encoder.

\begin{equation}
    H_x = \text{BERT}(S_x)
\end{equation}
\begin{equation}
    H_d = \text{BERT}(S_d)
\end{equation}
In order to avoid the requirement of a reference summary, similar to \cite{sun2019feasibility}, we measure the semantic quality of the target summary $x$ by calculating the semantic similarity between $x$ and its source document $d$. Thus the semantic quality score is:

\begin{equation}
   S\_Score(x) = Sim(H_d^0, H_x^0),
\end{equation}
where $Sim$ refers to cosine similarity, $H^0$ denotes the hidden state corresponding to token\texttt{[CLS]}.

\subsection{Evaluating linguistic quality}
\label{section:evaluate-language-quality}
For a summary $x$ and its sequence of tokens $S_x$, the exact operations to obtain its linguistic quality score are as follows.

We first use the BERT encoder to get the representation of the summary $x$. \begin{equation}
    H_x = \text{BERT}(S_x),
\end{equation}
where $H_x \in \mathbb{R}^{N \times K}$, $N$ is the sequence length and $K$ means the hidden size of the BERT encoder. Then we calculate the probability of the sequence based on this representation.

\begin{equation}
    P_x = \text{softmax}\bigg(W_1^\top\big(\sigma(W_0^\top H_x)\big))\bigg),
\end{equation}

where $W_0 \in \mathbb{R}^{K \times K}$ and $W_1 \in \mathbb{R}^{K \times V}$ denotes two weight parameters and we omit biases here. $V$ stands for the vocabulary size. $\sigma$ is an activation function, which is GELU in our experiments.  A softmax operation is applied to every token's embeddings to predict a probability distribution at each position in the sequence. Here we use $p_x^i$ to represent the probability of the $i$-th token to be the same as $S_x^i$. Motivated by the perplexity, the linguistic quality of $x$ can be calculated as:

\begin{equation}
   L\_Score(x) = \frac{1}{|x|}\sum_i^n{log~p_x^i}
\end{equation}

\begin{table*}[ht]
\centering

\scalebox{1}{
\begin{tabular}{p{0.9\linewidth}} 
\toprule

        \textbf{Original summary:}\\
        Kristina Patrick from Alaska filmed her German Shepherd Pakak performing a very skillful trick. Footage shows the pup taking the ball from her mouth with her paws and holding it up high in the air to admire it. She then carefully lowers it back down to the starting point. \\
        \cmidrule(l){1-1}
        \textbf{Negative samples:}\\
        \underline{1. delete words}\\
        Patrick $\land$ from Alaska filmed her German Shepherd Pakak performing a very skillful trick. Footage shows the pup taking the $\land$ from her $\land$ with her paws and holding it up high in the air to $\land$ it. She then carefully lowers it back down to the starting point.\\
        \underline{2. add sentences}\\
        Kristina Patrick from Alaska filmed her German Shepherd Pakak performing a very skillful trick. Footage shows the pup taking the ball from her mouth with her paws and holding it up high in the air to admire it. She then carefully lowers it back down to the starting point. \sout{PAKAK 's owner says she loves playing with balls.}\\
        \underline{3. disorder words}\\
        Kristina Patrick skillful Alaska filmed her performing Shepherd a German Pakak very from trick. Footage shows the pup taking the ball from admire mouth with and paws her holding it up high her to air the in it. She then back lowers it carefully to down the starting point.\\
\bottomrule
\end{tabular}
}
\caption{An example of negative sampling.}
\label{tab:neg-sampling}
\end{table*}

\subsection{Evaluating Both Dimensions}
\label{section:evaluate-both-dims}
In order to capture both the linguistic and semantic aspects, we develop our final metric by linearly combining the S\_Score and L\_Score. We call it LS\_Score, which is a trade-off between the semantic score and linguistic score. 

\begin{equation}
\resizebox{.89\linewidth}{!}{$
    \displaystyle
   LS\_Score(x) = \alpha L\_Score(x) + \beta S\_Score(x)
$}
\label{eq:LS-eval}
\end{equation}

The $\alpha$ and $\beta$ are used to scale the L\_Score and the S\_Score. In our experiments we fix $\alpha=0.01$ and $\beta=1$ to scale the L\_Score and the S\_Score.

\subsection{Contrastive Training}
To alleviate the requirement of reference summaries as well as given human evaluation scores, we develop a new unsupervised training framework via contrastive learning. Intuitively, for a given good summary, if we make some noise, e.g. disordering the words/sentences, we can easily create a summary with worse quality. Then we can compare these two summaries to get a contrastive loss. In practice, we can use human generated summaries in the training data as the "good" summaries, however, they can also be replaced with other good machine-generated summaries. We do not require any reference summaries in the test phase, i.e. for a candidate summary without known reference summaries we can also predict a score for it. That increases the flexibility and generalizability of our evaluation method.

Given a base summary $r$, assume we make some noise and get a set of negative samples $\hat{X_r}$, we formulate a ranking loss function as follows:

\begin{equation}
\resizebox{.89\linewidth}{!}{$
    \displaystyle
 Loss = \sum_{r \in \mathcal{R}} \sum_{\hat{x} \in \hat{X}_r} max(0,  1-(LS\_Score(r) - LS\_Score(\hat{x})))
$}
\label{formula:objective-func-S}
\end{equation}
where $\mathcal{R}$ denotes the set of original summaries in the training set and $\hat{X}_r$ is the set of corresponding noisy variants of a training sample $r$. For a batch of $(r, \hat{X_r})$, we obtain their scores predicted by an evaluation model and then update model parameters (including fine-tuning BERT) using the gradients of the loss function . In this way, we train the model to better distinguish between good and bad summaries. 

Since we evaluate the summaries from two different aspects, for each aspect we create different types of noisy samples. For semantic quality, one straightforward strategy is to randomly remove some words or sentences in the original summary to get a new negative sample. Obviously the created new summary will encounter information loss compared to the original one, so its evaluator score will be lower. In our experiments, we randomly select 20\% words (with no consideration of word types) to delete. We do not delete entire sentences because most of the summaries have only very few sentences (as shown in Table~\ref{tab:dataset statistics}, the average number of sentences in a reference is 1.43 and 3.88 in Newsroom and CNN/Daily Mail, respectively), thus deleting sentences will cause too much information loss, which doesn't benefit the model's ability to distinguish good from bad.  

In addition, we do not want the generated summaries to have too much redundant information. So we create another type of negative samples by adding redundant sentences. The redundant sentences are chosen randomly from the original document. Firstly we extract sentences from the original document. Then we filter out the sentences that are most similar to each sentence in the reference. At last, we randomly sample the redundant sentences from the remaining sentences in the reference.

For linguistic quality, the negative samples can be generated by either disordering the words/sentences or deleting words. Both of the operations will lead to loss of coherence or fluency. So the negative sampling strategy in this case is as follows: 1) randomly rotating the order of sentences or the order of words within a sentence. 2) randomly deleting some of the words in the original summary. Note that the second strategy is also used in generating noisy samples for semantic quality, but our LS\_Score combines both semantic and linguistic quality, so we do not explicitly discriminate the two aspects for this type of negative samples.

Table~\ref{tab:neg-sampling} shows three examples of our negative samples, each of which represents one type of negative samples respectively. By differentiating the original summaries and the negative samples we enforce our evaluator to capture various aspects of the summary quality. The trained evaluator can then be used for evaluating summaries with unknown references. In our experiments, we generate only one negative sample per type of operations for each base summary, i.e. each base summary has 3 negative samples.

\section{Experiments}
We conduct our experiments to answer the following questions:
\begin{itemize}
    \item Does our contrastive learning method obtain better performance over other baselines even without reference summaries?
    \item Can our evaluator capture the expected aspects of summary qualities, and does it outperform others under the same contrastive learning framework?
    \item Is our method generalizable to different datasets? That is, how does it perform if we train the metric on one dataset and test on another one?
\end{itemize}

\subsection{Experimental Settings}
The encoder in our experiments to convert token sequence into embeddings is BERT \cite{devlin2019bert}. We simply use a pretrained BERT model \texttt{bert-base-uncase} which has 12 layers, a hidden size of 768, 12 attention heads and 110M parameters in total.\footnote{\url{https://github.com/google-research/bert}} Our model is implemented based on the HuggingFace Transformers.\footnote{\url{https://github.com/huggingface/transformers}} The max length of sequence we use for BERT encoding is 512, so we truncate the sequence longer than 510 tokens (despite the special tokens \texttt{[CLS]} and \texttt{[SEP]}).\footnote{Our code is publicly available at \url{https://github.com/whl97/LS-Score.git}}.

\begin{table}[ht]
\centering
\scalebox{0.95}{
\begin{tabular}{lcc} 
\toprule

    \textbf{ } & \textbf{Newsroom} & \textbf{CNN/Daily} \\
    \cmidrule(l){1-3}
    \textbf{\# of doc-ref pairs} & 108,802  & 10,932   \\
    \textbf{\# of sens in doc} & 31.08  & 34.20   \\
    \textbf{\# of words in doc} & 861.90   & 882.25   \\
    \textbf{\# of sens in ref} & 1.43   & 3.88   \\
    \textbf{\# of words in ref} & 34.90   & 64.87  \\
    \textbf{\# of systems} &7  &4\\
    \textbf{\# of generated sums} &420  &1996\\

\bottomrule
\end{tabular}
}
\caption{Datasets statistics}
\label{tab:dataset statistics}
\end{table}

\subsection{Datasets}

We conduct empirical studies on two benchmark single-document summarization datasets. These datasets both have original documents, their corresponding human-authored summaries (i.e. references) and also some model-generated summaries that are manually rated in several dimensions, so we can compare different evaluation methods by their correlation with human ratings.

\textbf{Newsroom}. Proposed by \citeauthor{grusky2018newsroom} (\citeyear{grusky2018newsroom}), this summarization dataset includes 1.3 million documents and human-written summaries. In this corpus, there are only 420 summaries with human ratings. These summaries are generated by 7 different extractive or abstractive summarization systems. Each document-summary pair is evaluated by three human raters in four dimensions (\texttt{coherence}, \texttt{fluency}, \texttt{informativeness}, and \texttt{relevance}). We take the mean score of three raters as the groundtruth human score for each summary. We use these summaries with human ratings as our test data. In order to prevent information leakage in the training process, we select our training data (108,802 document-reference pairs) with no overlapped reference summaries with the test data. It means we do not use any reference summaries in our test data for training. The data statistics are shown in Table~\ref{tab:dataset statistics}.

\textbf{CNN/Daily Mail}. This dataset was first proposed by \citeauthor{hermann2015teaching} (\citeyear{hermann2015teaching}) using news documents for question answering research and was subsequently extended to the area of summarization by \citeauthor{nallapati2016abstractive} (\citeyear{nallapati2016abstractive}). \citeauthor{chaganty2018price} (\citeyear{chaganty2018price}) provided human scores for 2,513 
references and system-generated summaries in three dimensions (\texttt{overall}, \texttt{fluency} and \texttt{redundancy}). We use 1,996 summaries generated by 4 systems for testing and 10,932 document-reference pairs for training. Similarly, there is no overlap of reference summaries between the training data and test data. Table~\ref{tab:dataset statistics} shows the data statistics of the training data. 

For both datasets, in the training data, we randomly selected 95\% of sentence-pairs for training and the remaining 5\% for validation.

\subsection{Baselines}
We adopt the following metrics as our baselines. Since this paper focuses on unsupervised approaches, we do not compare with the metrics training with human ratings. 

\textbf{ROUGE}. This metric has been the most frequently used automatic metric for summarization evaluation. It evaluates the quality of a summary by comparing it to a human-authored reference. The essence of the comparison is to measure the overlapping units (such as n-gram or word sequences) between the summary and the reference~\cite{lin2004looking}. 

\textbf{METEOR}. Proposed by \citeauthor{banerjee2005meteor} (\citeyear{banerjee2005meteor}), this metric evaluates a candidate string by measuring the harmonic mean of unigram-precision and unigram-recall between the candidate string and a reference string.

\textbf{BERTScore}. This metric was proposed by \citeauthor{zhang2019bertscore} (\citeyear{zhang2019bertscore}), it utilizes token-level contextual embeddings generated by a pretrained language model (here we use BERT). The evaluation score is calculated by computing similarity between the embeddings of the summary to evaluate and the reference. The BERTScore includes three metrics $R$ (recall), $P$ (precision)and $F$ (F1 score).

\textbf{WMS/SMS/S+WMS}. \citeauthor{10.5555/3045118.3045221} (\citeyear{10.5555/3045118.3045221}) proposed word mover’s distance (WMD) to calculate the minimum cost of moving a sequence into the other. They treat each sequence as a bag of words and each word is represented by its word embeddings. The WMD can then be transformed into a similarity (WMS) \cite{clarketal2019sentence}. On the basis of \textbf{WMS}, \cite{clarketal2019sentence} (\citeyear{clarketal2019sentence}) designed to measure the similarity of two sequences by calculating sentence mover’s distance to enhance the ability of evaluating multi-sentence texts. They introduced two metrics: sentence mover’s distance (\textbf{SMS}) and sentence and word mover’s distance (\textbf{S+WMS}). \textbf{SMS} uses sentence embeddings instead of word embeddings and represents each sequence as a bag of sentences and \textbf{S+WMS} combines both sentence and word embeddings and represents each sequence as a bag of both sentences and words.

\textbf{MoverScore}. Also inspired by WMD,  \citeauthor{zhao2019moverscore} (\citeyear{zhao2019moverscore})  represented both the reference and the candidate text as a sequence of n-gram embeddings and calculate the WMD between two sequences. We report the result of the best models described in their paper that use a BERT pretrained on MNLI dataset to generate the n-gram embeddings and PMeans as the aggregator.

\textbf{BERT+Cos+Ref}. This metric uses BERT as the encoder and calculates the cosine similarity between the embeddings of the reference and the candidate summary.

\textbf{BERT+Cos+Doc}. This metric is similar to \textbf{BERT+Cos+Ref}, but it measures the similarity between the source document and the candidate summary. This is the only \emph{reference-free metric} in the baselines.

\begin{table}[ht]
\centering
\scalebox{0.80}{

\begin{tabular}{lcccc}
\toprule
& \textbf{Coh.} & \textbf{Flu.} & \textbf{Inf.} & \textbf{Rel.} \\
\cmidrule(l){1-5}
\textbf{ROUGE-1} 	&0.2446	&0.1991	&0.3371	&0.3028 \\
\textbf{ROUGE-2}	   &0.1133 &0.0763	&0.1816 	&0.1385 \\
\textbf{ROUGE-L}	  &0.2164	&0.1736 	&0.3178	&0.2700 \\
\textbf{METEOR}  &0.3325 &0.3347	&0.4424 	&0.4117 \\
\textbf{BERTScore-R}	 &0.2355	&0.2227 	&0.2972 	&0.2787 \\
\textbf{BERTScore-P}	 &-0.0263	&-0.0221	&-0.0215	&-0.0302 \\
\textbf{BERTScore-F} 	&0.1206 	&0.1072 	&0.1681 	&0.1426 \\
\textbf{WMS}	&0.2389 	&0.2355 	&0.3003 	&0.2406 \\
\textbf{SMS}	&0.2394 	&0.2400 	&0.2946 	&0.2401 \\
\textbf{S+WMS}	&0.2433 	&0.2405 	&0.3022 	&0.2432 \\
\textbf{MoverScore}   &0.1458   &0.1021    &0.2070    &0.1724 \\
\textbf{BERT+Cos+Ref}	&0.0452 	&0.0333 	&0.0475 	&0.0534\\
\textbf{BERT+Cos+Doc}	&0.3998 	&0.3492 	&0.4530 	&0.4279 \\
  \cmidrule(l){1-5}
\textbf{LS\_Score}	&\textbf{0.6390} 	&\textbf{0.5933} 	&\textbf{0.7163} 	&\textbf{0.6563}  \\
\bottomrule
\end{tabular}
}
\caption{Spearman correlation w.r.t. coherence (Coh.), fluency (Flu.), informativeness (Inf.) and relevancy (Rel.) on Newsroom. Best results are in bold.}
\label{tab:corrlation-newsroom}
\end{table}

\begin{table}[ht]
\centering

\scalebox{0.8}{
\begin{tabular}{lccc}
\toprule
  &\textbf{Overall}	&\textbf{Grammar} &\textbf{Redundancy}\\
  \cmidrule(l){1-4}
\textbf{ROUGE-1}	&0.1953	&0.0975	&0.2174\\
\textbf{ROUGE-2}	&0.1355	&0.0701	&0.1442\\
\textbf{ROUGE-L}	&0.1925	&0.0973	&0.2072\\
\textbf{METEOR}	&0.0773	&0.0173	&0.1147\\
\textbf{BERTScore-R}	&0.2628	&0.1721	&0.2780\\
\textbf{BERTScore-P}	&0.1754	&0.1828	&0.1180\\
\textbf{BERTScore-F}	&0.2536	&0.2041	&0.2348\\
\textbf{WMS}	&0.1809 	&0.1080 	&0.2274 \\
\textbf{SMS}	&0.1814 	&0.1021 	&0.2313 \\
\textbf{S+WMS}	&0.1830 	&0.1075 	&0.2314 \\
\textbf{MoverScore} &0.2220 &0.1522 &0.2289 \\  
\textbf{BERT+Cos+Doc}	&0.1484 	&0.1110 	&0.1237 \\
\textbf{BERT+Cos+Ref}	&0.2130 	&0.1316 	&0.2284 \\
  \cmidrule(l){1-4}
\textbf{LS\_Score}	&\textbf{0.3342} 	&\textbf{0.2664} 	&\textbf{0.2875}\\ 
\bottomrule
\end{tabular}
}
\caption{Spearman correlation on CNN/Daily Mail.}
\label{tab:corrlation-cnndm}
\end{table}

\subsection{Experiment Results}
The usual practice of evaluating a summarization evaluation metric is to measure its average summary-level correlation with human judgements, i.e. to measure the correlation between the predicted scores and the human scores across all the test summaries. We evaluate our methods on the aforementioned two datasets. We implemented our final model ( LS\_Score with contrastive learning), as we introduced in \ref{section:evaluate-both-dims}. For each dataset, we train our models on the document-reference pairs in the training data, and test on the machine-generated summaries without comparing with reference summaries.

\subsubsection{Comparison with Other Methods}
The Spearman correlations between different evaluation methods and human evaluation in four dimensions on Newsroom are shown in Table~\ref{tab:corrlation-newsroom}. Even though most of baselines are with reference summaries, our reference-free evaluator (LS\_Score) still achieves best correlations in all of the different dimensions. By capturing both the semantic quality and semantic quality in the evaluator's scoring function as well as our negative sampling strategies, our method outperforms other previous metrics a lot in both linguistic dimensions (\texttt{coherence}, \texttt{fluency}) and semantic dimensions (\texttt{informativeness}, \texttt{relevancy}). Especially, it is also superior to another unsupervised reference-free method, BERT+Cos+Doc.

Furthermore, we observe that BERT+Cos+Doc achieves a better overall performance on Newsroom as compared to BERT+Cos+Ref. This is probably due to the short lengths of the summaries on the Newsroom dataset (mostly one sentence). A possible explanation is that the short reference summaries fail to capture all the important information of original documents. As a result, directly comparing with document representations will suffer much less information loss.  

Table \ref{tab:corrlation-cnndm} shows the Spearman correlations on CNN/Daily Mail. As mentioned before, this dataset focuses more on evaluating the linguistic quality of summaries. One interesting comparison is between our model and BERTScore-R. On \texttt{redundancy} BERTScore-R is comparable but its \texttt{grammar} ratings is much worse than ours, which also leads to a worse overall performance.

\subsubsection{Ablation Study for Evaluator Selection}
We further conduct experiments to show the benefit of using our evaluator. A commonly used BERT-based evaluator is to add a linear regressor to the BERT representations \cite{Xenouleas2019SumQE}. We implement an evaluator (called BERT+Linear) that also uses a linear regressor to map the BERT embeddings of summaries into a score. We train this evaluator under our contrastive learning framework with the same negative samples, and compare its results with ours. Table~\ref{tab:ablation-newsroom} and Table~\ref{tab:ablation-cnndm} show the comparison results, and our model is superior to BERT+Linear a lot in most cases. One thing worth mentioning is that this ablation model already obtained better results than most of the baselines in Table~\ref{tab:corrlation-newsroom} and Table~\ref{tab:corrlation-cnndm}, which further demonstrate the power of our contrastive learning framework.

\begin{table}[ht]
\centering
\scalebox{0.8}{
\begin{tabular}{lcccc}
\toprule
& \textbf{Coh.} & \textbf{Flu.} & \textbf{Inf.} & \textbf{Rel.} \\
\cmidrule(l){1-5}
\textbf{Bert+Linear} 	&0.4213 	&0.4511 	&0.3075 	&0.3400  \\

\textbf{LS\_Score}	&\textbf{0.6390} 	&\textbf{0.5933} 	&\textbf{0.7163} 	&\textbf{0.6563}  \\
\bottomrule
\end{tabular}
}
\caption{Ablation studies on Newsroom. The models use the same contrastive learning framework but different evaluators.}
\label{tab:ablation-newsroom}
\end{table}

\begin{table}[ht]
\centering

\scalebox{0.80}{
\begin{tabular}{lccc}
\toprule
  &\textbf{Overall}	&\textbf{Grammar} &\textbf{Redundancy}\\
  \cmidrule(l){1-4}
\textbf{Bert+Linear}	&0.2711 	&\textbf{0.2886} 	&0.1664 \\

\textbf{LS\_Score}	&\textbf{0.3342} 	&0.2664 	&\textbf{0.2875}\\ 
\bottomrule
\end{tabular}
}
\caption{Ablation studies on CNN/Daily Mail. The models use the same contrastive learning framework but different evaluators.}
\label{tab:ablation-cnndm}
\end{table}

\subsubsection{Cross-dataset Transferability}
Although the generated summaries are from documents not included in the training data,  we still do experiments to further verify the transferability of our methods by training on one dataset and testing on the other dataset's test data. The performance of our method trained on CNN/Daily Mail and tested on Newsroom is shown in Table \ref{tab:corrlation-newsroom-transferability}, and the one trained on Newsroom and tested on CNN/Daily Mail are presented in Table \ref{tab:corrlation-cnndm-transferability}. We call this model LS\_Score\_cross. For easy comparison, we also take some values in Table~\ref{tab:corrlation-newsroom} and Table~\ref{tab:corrlation-cnndm}.
As shown in Table \ref{tab:corrlation-newsroom-transferability} and \ref{tab:corrlation-cnndm-transferability}, the cross-data training makes the performance of LS\_Score\_cross slightly lower than the original LS\_Score in most cases, but it still outperform all other baselines. This shows that our evaluation method is very flexible to be used. Even trained on different datasets, it can still achieve very good results. 

\begin{table}[ht]
\centering
\scalebox{0.80}{
\begin{tabular}{lcccc}
\toprule
& \textbf{Coh.} & \textbf{Flu.} & \textbf{Inf.} & \textbf{Rel.} \\
\cmidrule(l){1-5}
\textbf{ROUGE-1} 	&0.2446	&0.1991	&0.3371	&0.3028 \\
\textbf{ROUGE-L}	  &0.2164	&0.1736 	&0.3178	&0.2700 \\
\textbf{BERTScore-R}	 &0.2355	&0.2227 	&0.2972 	&0.2787 \\
\textbf{MoverScore}   &0.1458   &0.1021    &0.2070    &0.1724 \\
\textbf{BERT+Cos+Doc}	&0.3998 	&0.3492 	&0.4530 	&0.4279 \\
  \cmidrule(l){1-5}
\textbf{LS\_Score}	&\textbf{0.6390} 	&\textbf{0.5933} 	&\textbf{0.7163} 	&\textbf{0.6563}  \\
\cmidrule(l){1-5}
\textbf{LS\_Score\_cross}	&\textit{0.6271} 	&\textit{0.5852} 	&\textit{0.7008} 	&\textit{0.6381}  \\
\bottomrule
\end{tabular}
}
\caption{Cross-dataset training results: Spearman correlation on Newsroom. The model of LS\_Score\_cross is trained on CNN/Daily Mail.}
\label{tab:corrlation-newsroom-transferability}
\end{table}

\begin{table}[ht]
\centering

\scalebox{0.8}{
\begin{tabular}{lccc}
\toprule
  &\textbf{Overall}	&\textbf{Grammar} &\textbf{Redundancy}\\
  \cmidrule(l){1-4}
\textbf{ROUGE-1}	&0.1953	&0.0975	&0.2174\\
\textbf{ROUGE-L}	&0.1925	&0.0973	&0.2072\\
\textbf{BERTScore-R}	&0.2628	&0.1721	&0.2780\\
\textbf{MoverScore} &0.2220 &0.1522 &0.2289 \\  
\textbf{BERT+Cos+Doc}	&0.1484 	&0.1110 	&0.1237 \\
  \cmidrule(l){1-4}
\textbf{LS\_Score}	&\textbf{0.3342} 	&\textbf{0.2664} 	&0.2875\\ 
  \cmidrule(l){1-4}
\textbf{LS\_Score\_cross}	&\textit{0.2874} 	&\textit{0.1915} 	&\textbf{\textit{0.2881}}\\ 
\bottomrule
\end{tabular}
}
\caption{Cross-dataset training results: Spearman correlation on CNN/Daily Mail. The model LS\_Score\_cross is trained on Newsroom.}
\label{tab:corrlation-cnndm-transferability}
\end{table}

\section{Conclusion}

In this paper, we propose a new evaluation method in the field of text summarization. We found that the quality of a summary can be evaluated in two separate dimensions: semantic quality and linguistic quality. Since human-authored references used in most of the existing metrics are costly, we investigate automatic evaluation metrics in an unsupervised reference-free setting. Leveraging powerful representations of BERT, our methods achieve the highest performance on two datasets. 
Although our experiments are only on single-document summarization datasets, our method can also be also extended to evaluation of multi-document summarization with slight changes, especially in the part of semantic quality evaluation.

\bibliography{emnlp2020}
\bibliographystyle{acl_natbib}

\end{document}